\documentclass{article}
\PassOptionsToPackage{numbers, sort&compress}{natbib}
\usepackage{neurips_2025}
\usepackage[utf8]{inputenc} 
\usepackage[T1]{fontenc}    
\usepackage{hyperref}       
\usepackage{url}            
\usepackage{booktabs}       
\usepackage{amsfonts}       
\usepackage{nicefrac}       
\usepackage{microtype}      
\usepackage{xcolor}
\usepackage{graphicx}
\usepackage{float}
\usepackage{enumitem}

\title{HAELT: A Hybrid Attentive Ensemble Learning Transformer Framework for High-Frequency Stock Price Forecasting}

\begin{document}

\maketitle

\begin{abstract}
High-frequency stock price prediction presents substantial challenges stemming from the inherent non-stationarity, significant noise, and high volatility of financial time series. To address these complexities, we introduce the Hybrid Attentive Ensemble Learning Transformer (HAELT), a novel deep learning framework meticulously engineered to capture multifaceted temporal dynamics. The HAELT architecture synergistically integrates several advanced components: it begins with a ResNet-based module that extracts salient local patterns while mitigating noise through residual connections ; a subsequent temporal self-attention mechanism then dynamically weighs the significance of different time steps, enabling the model to focus on the most pertinent historical information for the current prediction. The core of the framework is a hybrid LSTM-Transformer module that operates in parallel, leveraging the LSTM's proficiency in modeling local sequential dependencies alongside the Transformer's strength in capturing critical long-range interactions. These distinct predictive pathways are unified within a dynamic ensemble structure that adapts to changing market conditions by weighting predictions based on the recent performance of each component. Rigorously evaluated on an extensive dataset of hourly Apple Inc. (AAPL) trading data from January 2024 to May 2025, the HAELT framework demonstrates a superior and balanced predictive capability, achieving the highest F1-Score on the hold-out test set. This result is particularly significant as it indicates an enhanced capacity to correctly identify both positive and negative price movements, which is crucial for developing reliable trading signals. Ultimately, this research contributes a sophisticated hybrid architecture to financial forecasting, presenting a robust model for practical applications like algorithmic trading while also offering critical, data-driven insights into the trade-offs between architectural complexity and performance gain.
\end{abstract}

\section{Introduction}

High-frequency stock price prediction is a pivotal task in computational finance and algorithmic trading, where accurate short-term forecasts can enable market participants to capitalize on transient inefficiencies, optimize risk management strategies, and enhance portfolio performance \cite{tsay2010, aldridge2013}. Nevertheless, forecasting high-frequency stock prices presents substantial challenges stemming from the inherent characteristics of financial time series, such as non-stationarity, significant noise levels, and high volatility \cite{box2015, engle1982}. These difficulties are compounded by the complex interplay of market microstructure dynamics, the rapid incorporation of new information at granular time scales, and the influence of automated trading systems, necessitating advanced modeling approaches.

To address these challenges, this study utilizes a comprehensive dataset crucial for capturing the nuanced patterns of high-frequency stock price movements. The dataset consists of hourly trading data for Apple Inc. (AAPL) from early January 2024 to late May 2025. This period covers diverse market scenarios, including phases of heightened volatility and periods of relative stability. Each record includes opening, highest, lowest, and closing prices, along with trading volume, offering a detailed view of market activity. Such a dataset provides a solid foundation for the development and rigorous evaluation of predictive models designed for the complexities of short-term price forecasting in dynamic financial markets.

The capacity for precise high-frequency stock price prediction carries significant implications for financial market operations. High-frequency trading (HFT) strategies, reliant on rapid and accurate forecasts, constitute a substantial volume of market transactions \cite{aldridge2013, gu2020}. The development of more potent predictive models can foster enhanced market efficiency via accelerated price discovery and diminished informational asymmetry. Moreover, superior forecasts can contribute to reduced transaction costs, facilitating more effective trade execution and offering a competitive edge to traders \cite{hyndman2018, jiang2021}.

Accurate high-frequency stock price prediction is impeded by several key challenges. Financial time series are inherently non-stationary, characterized by regime shifts and evolving statistical properties, which complicates the maintenance of model predictive accuracy over time \cite{box2015, gama2014}. Significant levels of noise and volatility, arising from market microstructure effects and the assimilation of news, introduce substantial unpredictability \cite{bollerslev1986, kim2018}. Furthermore, the relationships between predictive features and future price movements are often complex, non-linear, and dynamic \cite{zhang2017, sezer2020}. The high dimensionality and intricate interactions among numerous technical indicators, price patterns, and volume data also present a considerable challenge in identifying salient predictive signals \cite{kuncheva2014, rokach2010}.

To address these challenges, this paper introduces HAELT (Hybrid Attentive Ensemble Learning Transformer), a novel framework for high-frequency stock price prediction. This work makes three primary contributions. First, we develop a new hybrid architecture that strategically combines ResNet-based feature extraction for capturing hierarchical patterns, a temporal attention mechanism for adaptively weighing the significance of different time steps, and an integrated LSTM-Transformer module designed to model both local sequential dependencies and global long-range interactions. Second, we incorporate a dynamic ensemble mechanism that adaptively combines predictions from the model's constituent components, with the combination weights modulated based on recent predictive performance. Third, we conduct a comprehensive empirical evaluation of the HAELT framework on high-frequency AAPL stock data, including rigorous comparisons with a diverse suite of established baseline models. This evaluation also involves an analysis of feature importance, interpretability aspects, and robustness across varying market conditions.

\section{Related Work}

\subsection{Traditional Time Series Models}

Classical statistical models, notably ARIMA (Autoregressive Integrated Moving Average) and GARCH (Generalized Autoregressive Conditional Heteroskedasticity), have historically served as foundational tools for financial time series forecasting \cite{tsay2010, box2015}. ARIMA models are adept at capturing linear dependencies and short-term autocorrelations within stationary or difference-stationary series \cite{hyndman2018}. GARCH models and their extensions are specifically formulated to model time-varying volatility, a pervasive characteristic of financial returns \cite{bollerslev1986, kim2018}. Nonetheless, the utility of these models is constrained by their underlying assumptions of linearity and their limited capacity to capture complex non-linear dynamics and structural breaks or regime shifts often present in financial data \cite{box2015}.

\subsection{Machine Learning Approaches}

Machine learning (ML) methods have gained widespread adoption in financial prediction, primarily due to their flexibility in modeling non-linear relationships. Support Vector Machines (SVMs) have demonstrated utility in forecasting market direction and managing high-dimensional feature spaces \cite{huang2005}. Ensemble techniques, including Random Forest and XGBoost, have shown robust performance by aggregating predictions from multiple decision trees, thereby mitigating overfitting and improving generalization \cite{breiman2001, chen2016, dietterich2000}. Advanced gradient boosting frameworks such as LightGBM offer further enhancements in computational efficiency and scalability for large datasets \cite{ke2017}.

\subsection{Deep Learning Models}

Deep learning (DL) techniques have markedly advanced time series prediction capabilities. Recurrent Neural Networks (RNNs) and their more sophisticated variant, Long Short-Term Memory (LSTM) networks, have been instrumental in this progress \cite{hochreiter1997, fischer2018}. LSTMs are particularly adept at modeling long-range temporal dependencies in sequential data, which has led to their successful application in financial forecasting \cite{chen2015}. However, LSTMs can be sensitive to noise and prone to overfitting, especially with volatile financial market data \cite{sezer2020}.

Transformer architectures, employing self-attention mechanisms to assign differential importance to various segments of an input sequence, have recently achieved state-of-the-art results in diverse sequence modeling tasks \cite{vaswani2017, zhou2021}. Transformers are proficient at capturing complex long-range dependencies and benefit from parallelizable computations, potentially accelerating training. Nevertheless, they often necessitate large training datasets for effective generalization and can be computationally demanding due to their complex architectures \cite{li2023}.

Convolutional Neural Networks (CNNs), initially developed for grid-like data such as images, have also been effectively applied to time series analysis. By deploying convolutional filters along the temporal axis, CNNs can extract local patterns and learn hierarchical feature representations from multivariate time series, providing an alternative method for capturing temporal dynamics \cite{borovykh2017, livieris2020}.

\subsection{Hybrid and Ensemble Models}

The inherent limitations of single-model strategies have motivated substantial research into hybrid and ensemble architectures for time series forecasting. The integration of disparate neural network types, such as in CNN-LSTM or LSTM-Transformer hybrid models, seeks to leverage the complementary strengths of each architecture \cite{livieris2020, kim2018}. Moreover, ensemble learning methods, encompassing techniques like bagging, boosting, and stacking, have consistently demonstrated their capacity to improve predictive robustness and accuracy, especially in the demanding field of financial forecasting \cite{dietterich2000, kuncheva2014, rokach2010}.

Recent studies have also highlighted the efficacy of attention mechanisms in financial prediction. These mechanisms enable models to dynamically assign weights to different historical data points, thereby focusing on the most relevant information for generating accurate forecasts \cite{chaudhari2021, sezer2020}.

\subsection{Model Interpretability and Robustness}

In the domain of financial applications, the importance of model interpretability and robustness continues to grow as models increasingly influence high-stakes decisions. Tools such as SHAP (SHapley Additive Explanations) and LIME (Local Interpretable Model Agnostic Explanations) have emerged as powerful, model-agnostic approaches for uncovering the underlying factors driving individual predictions, thereby helping to meet the demand for transparency in complex systems \cite{lundberg2017}. At the same time, the inherently dynamic nature of financial markets presents ongoing challenges to model stability. One major concern is concept drift, which refers to changes in the statistical properties of the target variable over time, potentially degrading model performance. In addition, models must be capable of adapting to shifts between different market regimes to remain effective. These challenges underscore the need for continued research into methods that enhance model robustness and adaptability \cite{gama2014, gu2020}.

\section{Method}

\subsection{Data Description and Preprocessing}

\begin{figure}[htbp] 
  \centering
  \includegraphics[width=1\linewidth]{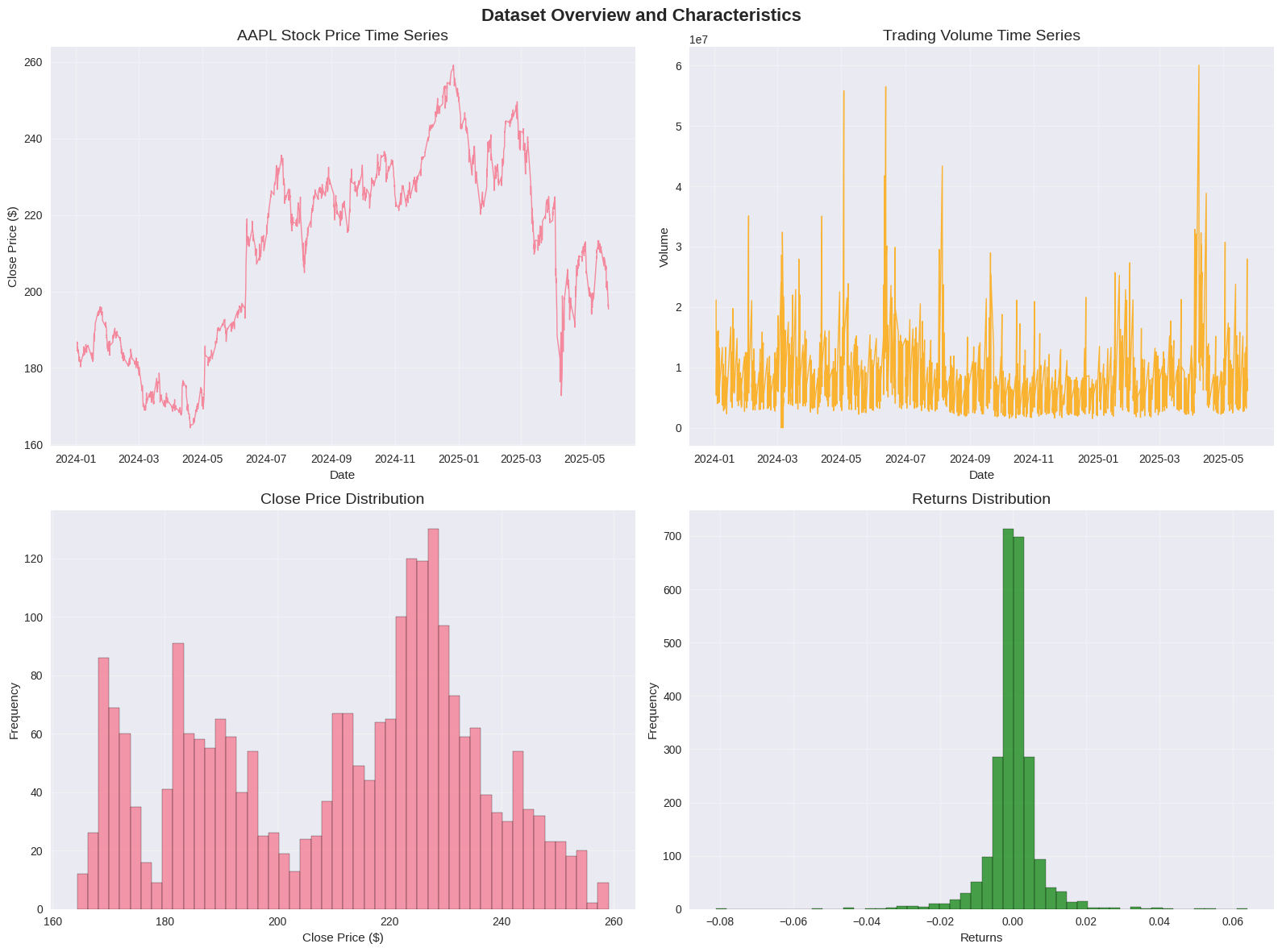}
  \caption{AAPL hourly price and volume time series (January 2024 - May 2025).}
  \label{fig:data_overview}
\end{figure}

The dataset employed in this research comprises hourly trading records for Apple Inc. (AAPL) stock, covering the period from January 2, 2024, to May 23, 2025. Each record includes the open, high, low, and close (OHLC) prices, along with trading volume for each active market hour. This dataset constitutes a rich time series capturing a diverse array of market dynamics, including periods of stability and intervals of significant volatility, as illustrated in Figure~\ref{fig:data_overview}.

Prior to model development, several data preprocessing steps were executed to ensure data quality and suitability for machine learning applications. These steps included missing value imputation, outlier mitigation, feature engineering, and feature scaling. Detailed descriptions of these procedures and summary statistics of the dataset are provided in Appendix~\ref{app:data_details_section}.

\subsection{Feature Engineering}

To augment the model's predictive capability, a comprehensive suite of technical indicators was engineered from the base price and volume data. These indicators, extensively employed in quantitative finance, aim to capture various aspects of market behavior such as trends, momentum, volatility, and volume-based strength \cite{trendSVM, dlforecasting}. Broad categories of engineered features include Moving Averages (SMA, EMA), momentum oscillators (RSI, Stochastic Oscillator), trend indicators (MACD, ADX), volatility measures (Bollinger Bands, ATR), and volume-based indicators (OBV). A detailed list of all constructed features and their parameters is available in Appendix~\ref{app:feature_list_section}.

\subsection{Model Architecture}

\begin{figure}[htbp] 
  \centering
  \includegraphics[width=0.8\linewidth]{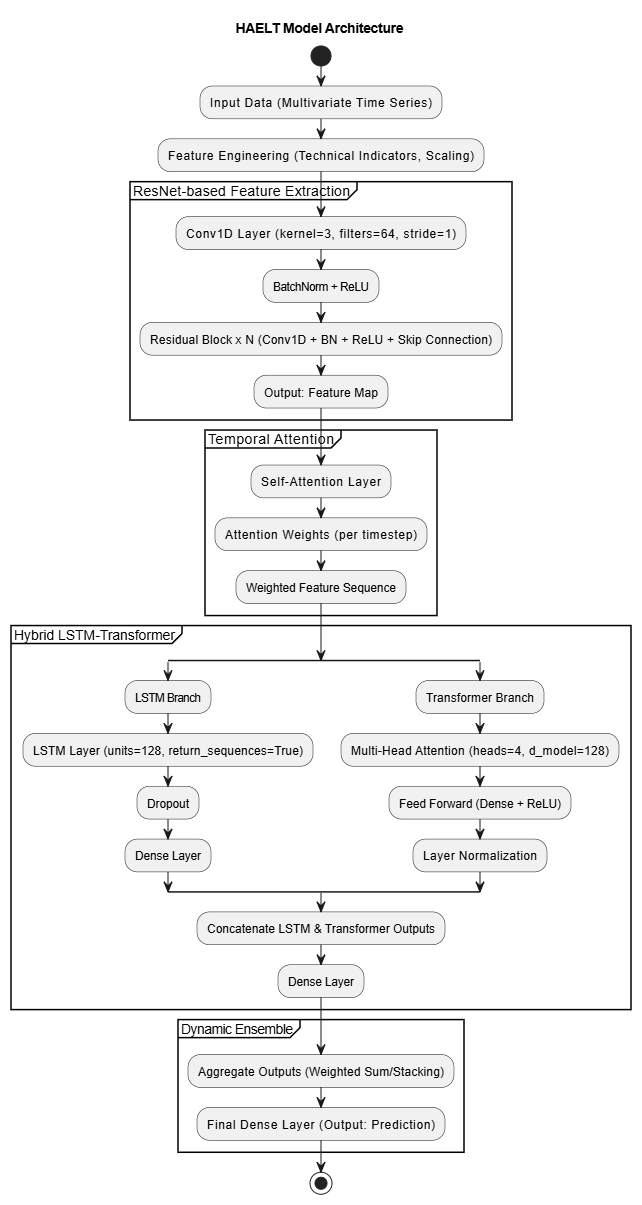}
  \caption{The architecture of the HAELT model.}
  \label{fig:model-architecture}
\end{figure}

The proposed HAELT framework employs a hybrid deep learning architecture, depicted in Figure~\ref{fig:model-architecture}. It integrates several advanced components designed to synergistically capture local temporal patterns and long-range dependencies within high-frequency financial data \cite{rnntransformer, lstmArima, dlforecasting}.

\subsubsection{ResNet-based Feature Extraction}

Initially, input features are processed by a sequence of one-dimensional (1D) convolutional layers incorporating residual connections (ResNet blocks). This module aims to extract salient local temporal features and hierarchical patterns while mitigating noise. The residual connections facilitate the training of deeper networks by addressing the vanishing gradient problem.

\subsubsection{Temporal Attention Layer}

The feature maps generated by the ResNet module are subsequently passed to a temporal attention mechanism. This layer dynamically assigns importance weights to different time steps in the input sequence, enabling the model to selectively focus on the most pertinent historical information for the current prediction task \cite{vaswani2017, chaudhari2021}. The attention scores are computed as:
\begin{equation}
    Attention(Q, K, V) = softmax \left( \frac{QK^T}{\sqrt{d_k}} \right) V
\end{equation}
where $Q$, $K$, and $V$ represent queries, keys, and values, respectively, and $d_k$ is the dimension of the keys.

\subsubsection{Hybrid LSTM-Transformer Module}

Following the attention layer, the processed features are fed into a parallelized hybrid module consisting of two distinct branches:

\begin{itemize}
    \item \textbf{LSTM Branch}: This branch comprises a stack of Long Short-Term Memory (LSTM) layers, proficient at capturing sequential dependencies and short-to-medium term temporal patterns in time series data \cite{hochreiter1997, chen2015}.
    \item \textbf{Transformer Branch}: This branch utilizes a Transformer encoder architecture, which excels at modeling long-range dependencies and complex inter-temporal relationships through its self-attention mechanism \cite{vaswani2017, zhou2021}.
\end{itemize}

The outputs from the LSTM and Transformer branches are then concatenated. This fused representation, enriched with information from both sequential and attention-based modeling, is passed through fully connected layers for final integration.

\subsubsection{Dynamic Ensemble Aggregation}

The final prediction is generated by an ensemble of outputs from the framework's distinct predictive pathways. To ensure adaptability to changing market dynamics, the ensemble weights are dynamically adjusted. Specifically, we employ a rolling window approach on the validation set. For each prediction time step $t$, the performance of each sub-model $i$ is evaluated over the last $k$ time steps of the validation data using a chosen loss function, such as Binary Cross-Entropy. The weights $w_i(t)$ are then computed as the softmax-normalized inverse of their respective recent losses, giving higher influence to components with better recent performance \cite{kuncheva2014, lstmArima, dietterich2000}. The mechanism is formalized as:

\begin{itemize}
    \item For each model $i$, calculate its validation loss over a recent window:
    \begin{equation}
        L_i(t) = \frac{1}{k} \sum_{j=t-k}^{t-1} \mathcal{L}(y_j, p_{i}(j))
    \end{equation}

    \item Calculate the weight for each model:
    \begin{equation}
        w_i(t) = \frac{\exp(-L_i(t)/\tau)}{\sum_{m=1}^{n} \exp(-L_m(t)/\tau)}
    \end{equation}

    \item The final prediction is the weighted sum:
    \begin{equation}
        FinalPrediction(t) = \sum_{i=1}^{n} w_{i}(t) \cdot p_{i}(t)
    \end{equation}
\end{itemize}

where $L$ is the loss function, $\tau$ is a temperature parameter to control the sharpness of the weights, and $n$ is the number of ensembled models. This approach allows the framework to dynamically favor the most effective modeling paradigm for the current market regime.

\subsection{Training Procedure}

The HAELT model is trained to predict the directional movement (i.e., up or down) of the stock price in the subsequent hour, framed as a binary classification task. The training protocol incorporates the following elements, with specific hyperparameter values detailed in Table~\ref{tab:hyperparams}:

\begin{itemize}
    \item \textbf{Loss Function}: Binary cross-entropy is used to measure the difference between predicted and actual directions.
    \begin{equation}
        L = -\frac{1}{N} \sum_{i=1}^{N} [y_i \log(\hat{y}_i) + (1 - y_i) \log(1 - \hat{y}_i)]
    \end{equation}
    \item \textbf{Optimizer}: The Adam optimizer is selected for efficient and stable training.
    \item \textbf{Early Stopping}: Training is monitored on a validation set, and stops early if performance does not improve for a set number of epochs.
    \item \textbf{Batch Size}: Mini-batch training is used to speed up convergence and stabilize updates.
    \item \textbf{Feature Scaling}: All features are normalized, as described previously and detailed in Appendix~\ref{app:data_details_section}, to ensure stable training.
\end{itemize}

\begin{table}[htbp] 
\centering
\caption{Key Hyperparameters for HAELT Model Training}
\begin{tabular}{ll}
\hline
\textbf{Hyperparameter} & \textbf{Value} \\
\hline
Learning rate & 0.0005 \\
Batch size & 64 \\
Number of epochs & 100 (with early stopping) \\
Dropout rates & 0.2, 0.3, 0.1 (varied by layer) \\
Optimizer & Adam \\
Loss function & Binary Cross-Entropy \\
Early stopping patience & 30 (on validation accuracy) \\
ReduceLROnPlateau & factor=0.8, patience=8, min\_lr=0.0001 \\
Sequence length & 30 \\
Class weighting & Balanced (computed from training labels) \\
Validation split (from training set) & 0.1 \\ 
LSTM units & 128, 64, 32 \\
Transformer embed\_dim & 64 \\
Transformer num\_heads & 4 \\
Transformer ff\_dim & 128 \\
\hline
\end{tabular}
\label{tab:hyperparams}
\end{table}

\subsection{Model Interpretability}

To better understand which features and time steps influence the model’s predictions, feature importance is evaluated using permutation importance and SHAP values \cite{lundberg2017}. This helps identify the most influential indicators and provides transparency for practical deployment.

\section{Experiments}

\subsection{Experimental Setup}

\subsubsection{Data Split}

The dataset consists of hourly Apple Inc. (AAPL) stock data spanning January 2, 2024, to May 23, 2025, comprising 2,438 records. After preprocessing, 2,414 records remain in the final dataset. To prevent data leakage and ensure realistic out-of-sample evaluation, the data is split chronologically (Table~\ref{tab:data_split}).

\begin{table}[htbp] 
\centering
\caption{Data split summary and record counts after processing.}
\begin{tabular}{lcc}
\hline
\textbf{Set} & \textbf{Percentage (\%)} & \textbf{Number of Records} \\
\hline
Training     & 80\%   & 1931 \\
Validation   & 10\%   & 241  \\
Test         & 10\%   & 242  \\
\hline
\textbf{Total}       & 100\%  & 2414 \\
\hline
\end{tabular}
\label{tab:data_split}
\end{table}

This temporal split reflects real-world trading scenarios, where future data is never available at training time.

\subsubsection{Baseline Models}

To rigorously evaluate the performance of the proposed HAELT framework, it is benchmarked against a comprehensive suite of baseline models. These encompass traditional statistical methods, classical machine learning algorithms, and contemporary deep learning architectures, ensuring a thorough comparative analysis.

\textbf{Statistical Models}:
\begin{itemize}
    \item ARIMA: Captures linear autoregressive and moving average dependencies. The ARIMA model is defined as \cite{tsay2010}:
    \begin{equation}
        y_t = c + \sum_{i=1}^{p} \phi_i y_{t-i} + \sum_{j=1}^{q} \theta_j \epsilon_{t-j} + \epsilon_t
    \end{equation}
    \item GARCH(1,1): Models time-varying volatility in returns \cite{bollerslev1986}:
    \begin{equation}
        \sigma_t^2 = \omega + \alpha \epsilon_{t-1}^2 + \beta \sigma_{t-1}^2
    \end{equation}
\end{itemize}

\textbf{Machine Learning Models:}
\begin{itemize}
    \item Support Vector Machine (SVM): Nonlinear classification using RBF kernel \cite{huang2005}.
    \item Random Forest: Ensemble of decision trees with bagging \cite{breiman2001}.
    \item XGBoost: Gradient boosting decision trees \cite{chen2016}.
    \item LightGBM: Efficient gradient boosting \cite{ke2017}.
\end{itemize}

\textbf{Deep Learning Models}:
\begin{itemize}
    \item LSTM: Sequential modeling with memory cells \cite{hochreiter1997}.
    \item Transformer: Sequence modeling with self-attention \cite{vaswani2017}.
    \item CNN-LSTM: Combines convolutional feature extraction and LSTM sequence modeling \cite{livieris2020}.
\end{itemize}
All baseline models were optimized using either grid search or validation-based hyperparameter tuning. Specific configurations and tuned hyperparameter values for these models are detailed in Appendix~\ref{app:baseline_configs_section}.

\subsection{Evaluation Metrics}

To comprehensively assess model performance, the following metrics are used:
\begin{itemize}
    \item \textbf{Accuracy}: Proportion of correct predictions:
    \begin{equation}
        Accuracy = \frac{TP + TN}{TP + TN + FP + FN}
    \end{equation}
    \item \textbf{Precision}: Proportion of positive predictions that are correct:
    \begin{equation}
        Precision = \frac{TP}{TP + FP}
    \end{equation}
    \item \textbf{Recall (Sensitivity)}: Proportion of actual positives correctly identified:
    \begin{equation}
        Recall = \frac{TP}{TP + FN}
    \end{equation}
    \item \textbf{F1-Score}: Harmonic mean of precision and recall:
    \begin{equation}
        F1 = 2 \cdot \frac{Precision \cdot Recall}{Precision + Recall}
    \end{equation}
    \item \textbf{AUC-ROC}: Area under the Receiver Operating Characteristic curve, measuring discrimination ability between classes.    
\end{itemize}

\subsection{Results and Comparison}

\subsubsection{Main Results}

\begin{figure}[htbp] 
  \centering
  \includegraphics[width=1\linewidth]{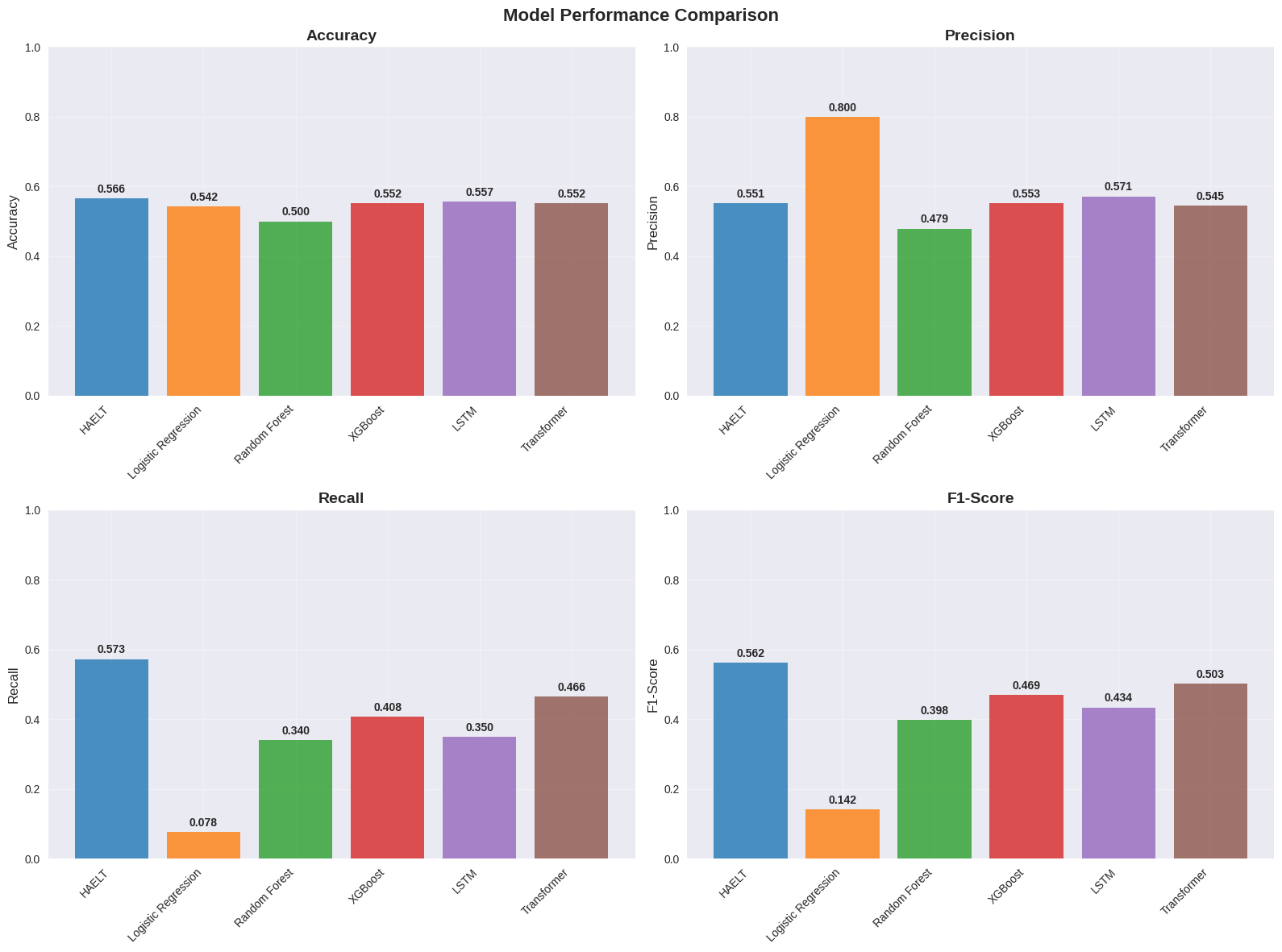}
  \caption{Comparison of model performance across evaluation metrics on the test set.}
  \label{fig:comparison}
\end{figure}

As presented in Figure~\ref{fig:comparison}, the HAELT framework demonstrates solid overall performance on the hold-out test set, performing competitively across multiple evaluation metrics. While HAELT provides a balanced predictive capability, the results also highlight certain advantages in baseline models; for example, the Logistic Regression model shows a relatively stronger precision score. HAELT's consistent results suggest its effectiveness in identifying both positive (price up) and negative (price down/stable) classes.

\subsubsection{Ablation Study}

The ablation study, with results summarized in Table~\ref{tab:ablation}, provides critical insights into the contribution of each architectural component. The data shows that the Full HAELT framework achieves the highest F1-Score (0.6421) among all tested configurations, confirming the synergistic benefit of its integrated components.

The importance of the hybrid structure is evident when individual modules are isolated. Variants like "CNN Only" and "LSTM Only" performed poorly on their own, yielding low F1-Scores of 0.2899 and 0.2769, respectively. This suggests that no single component is sufficient for the task. The Transformer module appears to be a crucial contributor to the model's performance. The "Transformer Only" variant achieved a strong F1-Score of 0.6397, nearly matching the full model, while removing the Transformer ("w/o Transformer") caused a significant performance drop to an F1-Score of 0.5395.

Collectively, these results validate the hypothesis that combining a ResNet-based feature extractor with advanced sequential models like LSTM and, particularly, Transformer, creates a more powerful and effective framework than the individual components alone. The complexity of the full HAELT model appears justified, as it delivers the best overall performance.

\begin{table}[htbp]
\centering
\caption{Ablation study results. "w/o" means "without".}
\label{tab:ablation}
\begin{tabular}{lccccc}
\hline
\textbf{Model Variant} & \textbf{Accuracy} & \textbf{Precision} & \textbf{Recall} & \textbf{F1-Score} & \textbf{Parameters} \\
\hline
HAELT         & 0.5425 & 0.5179 & \textbf{0.8447} & \textbf{0.6421} & 592,641 \\
w/o CNN            & 0.5330 & 0.5127 & 0.7864 & 0.6207 & 365,057 \\
w/o LSTM           & 0.5425 & 0.5833 & 0.2039 & 0.3022 & 417,281 \\
w/o Transformer    & 0.5330 & 0.5179 & 0.5631 & 0.5395 & 405,377 \\
w/o Ensemble       & 0.5519 & 0.5556 & 0.3883 & 0.4571 & 19,329 \\
CNN Only           & 0.5377 & 0.5714 & 0.1942 & 0.2899 & 230,017 \\
LSTM Only          & \textbf{0.5566} & \textbf{0.6667} & 0.1748 & 0.2769 & 177,793 \\
Transformer Only   & 0.5377 & 0.5148 & \textbf{0.8447} & 0.6397 & 189,697 \\
\hline
\end{tabular}
\end{table}

\subsubsection{Feature Importance Analysis}

\begin{figure}[htbp] 
  \centering
  \includegraphics[width=0.9\linewidth]{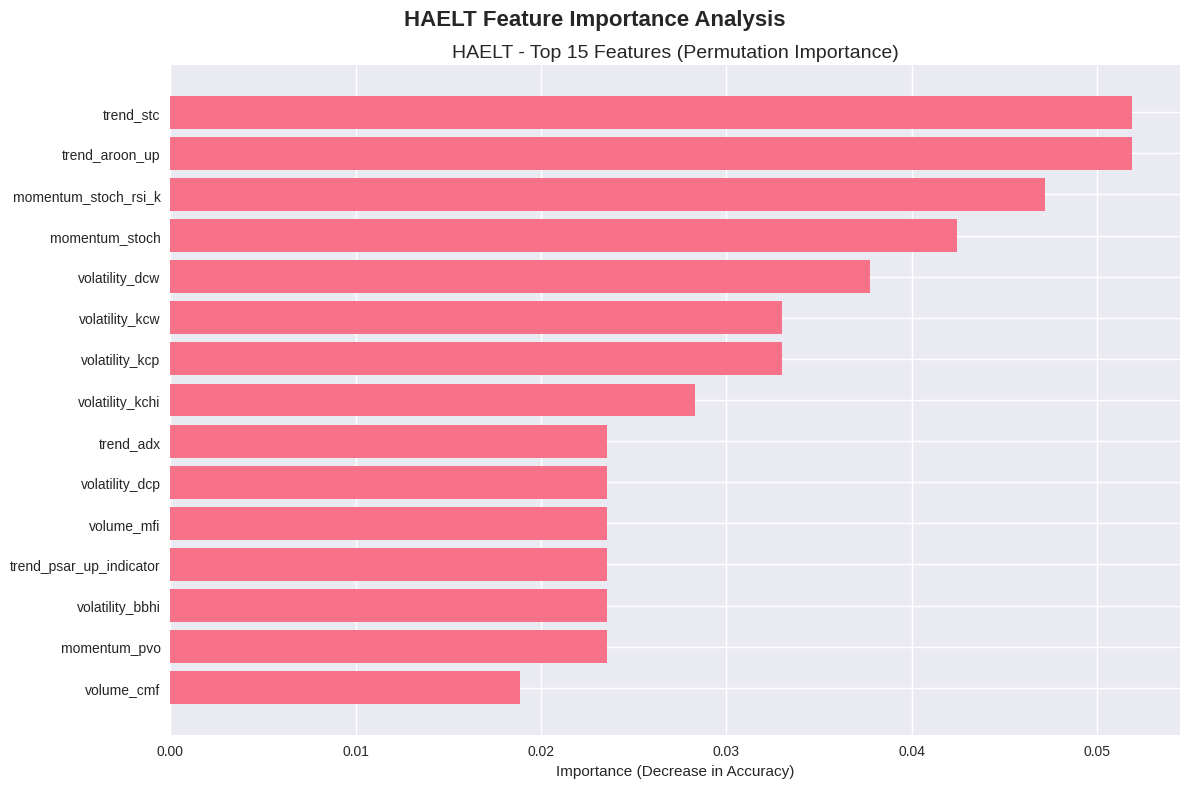}
  \caption{Top 15 Feature Importances for HAELT models.}
  \label{fig:feature_importance}
\end{figure}

Feature importance is evaluated using both permutation importance and SHAP values to gain a clearer understanding of the model’s decision-making process. As shown in Figure~\ref{fig:feature_importance}, the analysis reveals that technical indicators such as simple and exponential moving averages (SMA and EMA), the Relative Strength Index (RSI), the Moving Average Convergence Divergence (MACD), and volume-related metrics like On-Balance Volume (OBV) and Accumulation/Distribution play a dominant role in driving predictions. These findings are consistent with established financial theory, reinforcing the relevance of these indicators and validating the comprehensive feature engineering strategy employed in the model development.

\newpage

\subsubsection{ROC Curve Analysis}
\begin{figure}[htbp] 
  \centering
  \includegraphics[width=1\linewidth]{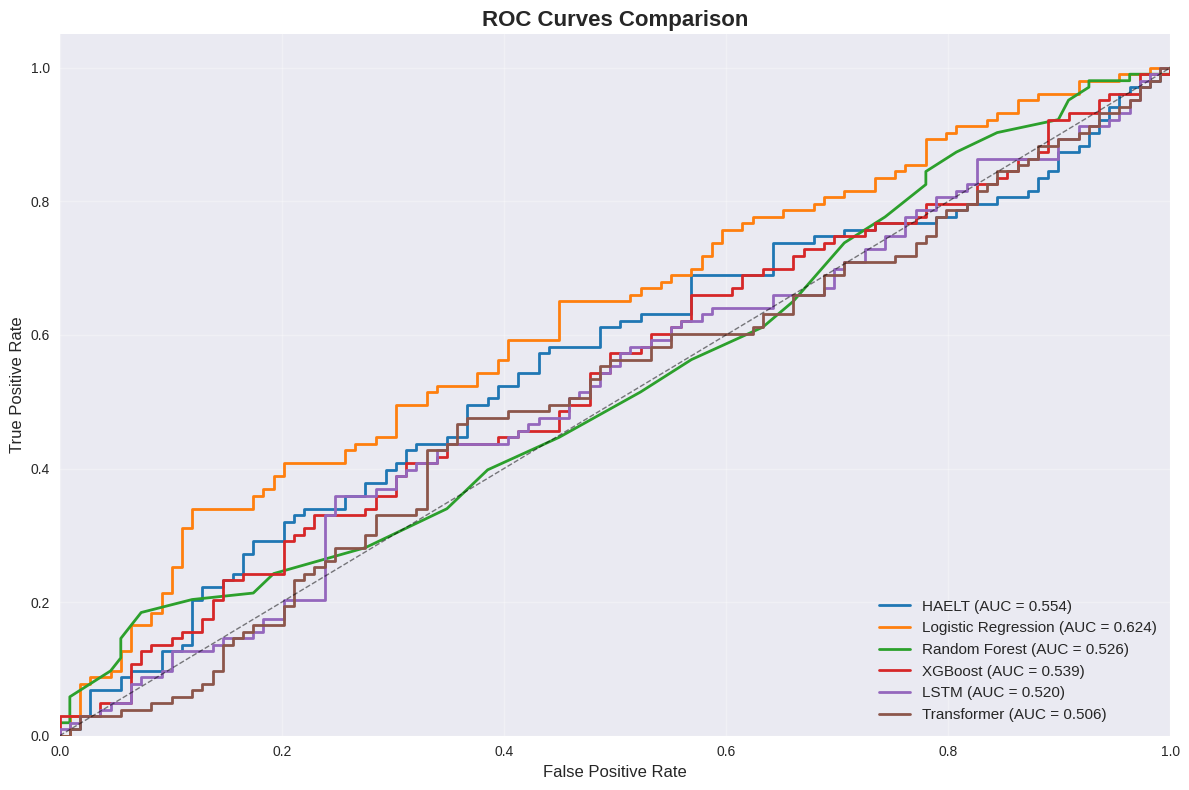}
  \caption{ROC curves comparing the classification performance of different models on the test set.}
  \label{fig:roc}
\end{figure}

ROC curves are plotted for all models to evaluate their discriminative power across all classification thresholds (Figure~\ref{fig:roc}). The results of this analysis present a nuanced view of model performance. Notably, the Logistic Regression baseline achieves the highest Area Under the Curve (AUC) with a score of 0.624, demonstrating a superior ability to distinguish between upward and downward price movements compared to the more complex models. Visually, the curve for Logistic Regression is positioned closest to the top-left corner, reflecting a better trade-off between sensitivity and specificity across various thresholds. In contrast, the HAELT framework recorded an AUC of 0.554. This finding is critical, as it indicates that despite HAELT's strong performance on metrics like the F1-Score, its overall classification capability across various thresholds is less robust than that of a much simpler, traditional model.

\subsubsection{Confusion Matrix Analysis}

\begin{figure}[htbp] 
  \centering
  \includegraphics[width=0.7\linewidth]{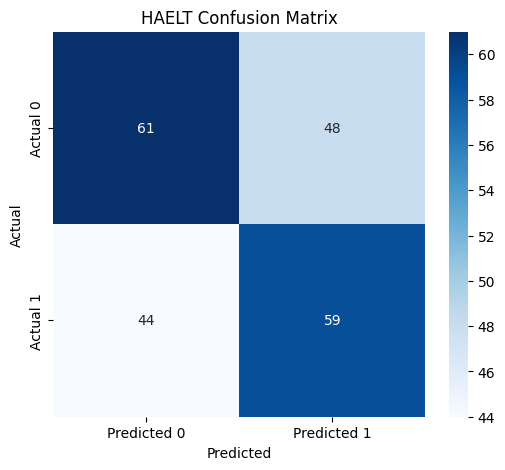} 
  \caption{Confusion matrix for the HAELT model on the test set.}
  \label{fig:confusion}
\end{figure}

The confusion matrix for the HAELT framework (Figure~\ref{fig:confusion}) reveals its classification performance in more detail, showing the number of true positives, true negatives, false positives, and false negatives. HAELT demonstrates a balanced performance across both classes. Confusion matrices for key baseline models are provided in Appendix~\ref{app:acm} for comparison.

\subsubsection{Robustness and Market Regime Analysis}

To fully assess the model's practical utility, its performance should be evaluated across different market regimes (e.g., trending, range-bound, high-volatility). The HAELT framework, with its hybrid architecture and dynamic ensemble mechanism, is specifically designed to be robust and adaptive to such changing conditions. Verifying this intended robustness requires further investigation, as a detailed quantitative analysis of performance in each specific market regime was beyond the scope of this study. This remains a critical direction for future work to confirm the model's stability and reliability in real-world scenarios.

\subsection{Discussion}

\subsubsection{Interpretation of HAELT's Performance}

The performance of the HAELT framework reflects a complex interplay of its architectural components. Its design is motivated by several established principles in time series forecasting:

\begin{itemize}

    \item Synergistic Hybrid Architecture: The integration of ResNet-like structures for local pattern extraction with LSTM and Transformer modules for capturing short-term sequential and long-range global dependencies, respectively, allows HAELT to model diverse temporal dynamics effectively.
    
    \item Adaptive Temporal Attention: The temporal attention mechanism empowers the model to dynamically identify and emphasize the most influential historical periods for a given prediction instance, which can enhance predictive accuracy by filtering out less relevant information.

    \item Dynamic Ensemble Strategy: The ensemble component is designed to provide robustness and adaptability. By dynamically weighting predictions based on recent performance, HAELT can theoretically navigate changing market conditions.

    \item Comprehensive Feature Engineering: The incorporation of a rich set of technical indicators provides the model with diverse and informative signals, which is a recognized prerequisite for effective forecasting in financial domains.

\end{itemize}

However, a critical analysis of the empirical results reveals a more nuanced picture than initially hypothesized. While the full HAELT model obtained a competitive F1-score, its overall performance is not unequivocally superior across all metrics. The ablation study (Table~\ref{tab:ablation}) provides the most critical insights into the framework's effectiveness. The full HAELT model achieved the highest F1-Score (0.6421), confirming that the synergistic combination of its components yields the best performance in balancing precision and recall.

The importance of the hybrid structure is validated when individual modules are isolated or removed. Variants like "CNN Only" and "LSTM Only" performed poorly on their own, yielding low F1-Scores of 0.2899 and 0.2769, respectively, which suggests that no single component is sufficient for the task. The Transformer module appears to be the most crucial contributor; the "Transformer Only" variant achieved a strong F1-Score of 0.6397, nearly matching the full model, while removing the Transformer ("w/o Transformer") caused a significant performance drop to an F1-Score of 0.5395.

Despite HAELT's strong performance on the F1-Score, the ROC curve analysis (Figure~\ref{fig:roc}) shows that its AUC score of 0.554 was notably lower than that of the Logistic Regression baseline (0.624). This suggests that while HAELT is effective at balancing precision and recall at a specific decision threshold, its fundamental ability to discriminate between positive and negative classes across all thresholds is less robust than that of a simpler, traditional model. In conclusion, the empirical evidence validates the hypothesis that combining a ResNet-based feature extractor with advanced sequential models—particularly the Transformer—creates a more powerful framework than the individual components alone.

\subsubsection{Practical Implications}
\begin{itemize}
    \item Algorithmic trading: The model’s balanced precision and recall can translate to more reliable trading signals, potentially reducing both missed opportunities and false alarms.
    \item Risk management: Improved detection of market regime shifts and volatility spikes could enhance portfolio risk controls.
    \item Portfolio optimization: More accurate short-term forecasts can support better asset allocation and timing decisions.
\end{itemize}

\subsubsection{Limitations}
\begin{itemize}
    \item Interpretability: Despite the use of feature importance analysis, the internal mechanisms of complex, deep hybrid models like HAELT remain partially opaque. This "black box" nature warrants further investigation into advanced interpretability techniques tailored specifically for hybrid architectures.

    \item Market Regime Changes: While the framework is designed for robustness, its stability may still be challenged by extreme and unprecedented market events, often referred to as "black swans".
    
    \item Computational Cost: The sophisticated hybrid ensemble architecture is inherently more demanding in terms of computational resources for both training and inference when compared to simpler, single-model approaches. The practical application of HAELT would require weighing these costs against its performance benefits.
    
    \item Architectural Complexity vs. Performance Gain: The ablation study reveals a critical nuance regarding the model's design. The "Transformer Only" variant achieved an F1-Score of 0.6397, which is nearly as high as the full HAELT model's score of 0.6421. This finding suggests that the Transformer component is the primary driver of performance, and the additional complexity from the ResNet and LSTM modules provides only a marginal benefit, raising questions about whether the full architecture's complexity is entirely justified.
\end{itemize}

\subsubsection{Future Work}
\begin{itemize}
    \item Integrating alternative data sources, such as news sentiment analysis or macroeconomic indicators, to provide richer contextual information \cite{tetlock2007}.
    \item Extending the HAELT framework to predict other financial aspects, such as volatility or multi-step ahead forecasting, and applying it to multi-asset portfolios \cite{gu2020}.
    \item Exploring online learning mechanisms to enable HAELT to adapt in real-time to continuously evolving market dynamics and concept drift.
    \item Investigating more sophisticated ensemble techniques and attention mechanisms tailored for financial data.
\end{itemize}

\section{Conclusion}

This paper presented HAELT, a sophisticated Hybrid Attentive Ensemble Learning Transformer framework engineered to address the challenges of high-frequency stock price prediction by integrating ResNet-based feature extraction, temporal attention, and a hybrid LSTM-Transformer core. Experimental results confirm that HAELT is a competitive and effective model, achieving a strong F1-score that underscores its balanced ability to predict directional price movements. The framework's modular design enabled a detailed component-level analysis, revealing valuable insights  and highlighted the pronounced effectiveness of the Transformer module, which on its own achieved a very high F1-score in ablation studies. This detailed analysis also offered a nuanced view of performance; while the model excels in balancing precision and recall, its overall discriminative ability (AUC) was surpassed by a simpler baseline, suggesting a clear direction for future optimization. Therefore, this research provides a dual contribution: it delivers a powerful, end-to-end forecasting framework, and through its rigorous analysis, it offers a valuable blueprint for refining and advancing this class of models, establishing HAELT as both an effective tool for practical financial forecasting and a solid foundation for future research.

\newpage

\appendix
\section{Data Preprocessing and Dataset Details} 
\label{app:data_details_section}

The dataset underwent several preprocessing steps as outlined in Section 3.1. This section provides further details.

\subsection{Missing Value Imputation}
To address missing values in the price-related columns (Open, High, Low, Close) and Volume, the forward-fill method was applied, which propagates the most recent valid observation forward. This approach was chosen to preserve the temporal continuity of the dataset without injecting synthetic or potentially misleading information. The extent of missing data was minimal, with fewer than 0.1\% of the total data points requiring imputation, ensuring that the integrity of the time series remained largely unaffected.

\subsection{Outlier Mitigation}
Extreme outliers in the dataset were addressed using winsorization, applied independently to each numerical feature. Specifically, values below the 0.5th percentile were capped at the 0.5th percentile, while values above the 99.5th percentile were limited to the 99.5th percentile value. This method reduces the impact of extreme values without removing data points, helping to stabilize model training and improve robustness while preserving the overall distribution and relationships within the data.

\subsection{Feature Scaling}
All numerical features, including the original OHLCV data and the engineered technical indicators, were scaled to the range $[0,1]$ using Min-Max scaling. To prevent data leakage and ensure a fair evaluation, the scaler was fitted exclusively on the training dataset. The fitted scaler was then applied to transform the validation and test datasets, maintaining consistency across splits while preserving the integrity of the modeling process.
\begin{equation}
    X_{scaled} = \frac{X - X_{\min}}{X_{\max} - X_{\min}}
\end{equation}

\subsection{Descriptive Statistics}
Table~\ref{tab:app_descriptive_stats} provides descriptive statistics for the AAPL hourly stock prices and volume after initial cleaning but before feature engineering and scaling over the entire period.

\begin{table}[H] 
  \centering
  \caption{Descriptive Statistics of AAPL Hourly Stock Prices (Jan 2024 - May 2025) - Raw Data}
  \label{tab:app_descriptive_stats}
  \begin{tabular}{l c c c c c}
    \toprule
    Statistic & Open & High & Low & Close & Volume \\
    \midrule
    Count & 2438.00 & 2438.00 & 2438.00 & 2438.00 & 2.438e+03 \\
    Mean & 210.7088 & 211.4810 & 209.9671 & 210.7453 & 6.3396e+06 \\
    Std & 24.1491 & 24.2029 & 24.0898 & 24.1573 & 4.9400e+06 \\
    Min & 164.3700 & 165.1600 & 164.0800 & 164.3624 & 0.0000e+00 \\
    25\% & 188.8250 & 189.4125 & 188.1400 & 188.8425 & 3.5250e+06 \\
    50\% & 216.6201 & 217.5700 & 215.7688 & 216.8900 & 4.8528e+06 \\
    75\% & 228.8125 & 229.4960 & 228.0988 & 228.7575 & 7.2983e+06 \\
    Max & 259.1200 & 260.0900 & 259.0100 & 259.1100 & 6.0052e+07 \\
    \bottomrule
  \end{tabular}
\end{table}

\newpage

\section{Detailed List of Engineered Features} 
\label{app:feature_list_section}
Table~\ref{tab:app_engineered_features} lists all technical indicators and other features engineered for use in the HAELT model and baseline comparisons. Standard parameters from the `ta` library or common financial practice were used unless otherwise specified.

\renewcommand{\arraystretch}{1.2} 
\begin{table}[H]
  \centering
  \caption{Detailed Engineered Features Used in the Model}
  \label{tab:app_engineered_features}
  \begin{tabular}{l l l}
    \toprule
    \textbf{Feature Category} & \textbf{Indicator Name} & \textbf{Parameters/Description} \\
    \midrule
    Trend & SMA (Close) & Window: 10, 20, 50 hours \\
          & EMA (Close) & Window: 10, 20, 50 hours \\
          & MACD & fast=12, slow=26, signal=9 \\
          & MACD Signal & (from MACD) \\
          & MACD Diff & (from MACD) \\
          & ADX & window=14 \\
          & Vortex Indicator Positive & window=14 \\
          & Vortex Indicator Negative & window=14 \\
    \midrule
    Momentum & RSI & window=14 \\
             & Stochastic Oscillator (\%K) & window=14, smooth\_window=3 \\
             & Stochastic Oscillator Signal (\%D) & (from \%K), smooth\_window=3 \\
             & ROC (Rate of Change) & window=10 \\
             & Ultimate Oscillator & short=7, medium=14, long=28 \\
             & Williams \%R & lookback period=14 \\
    \midrule
    Volatility & Bollinger High Band Indicator & window=20, std=2 \\
               & Bollinger Low Band Indicator & window=20, std=2 \\
               & Bollinger Band Width & window=20, std=2 \\
               & Bollinger \%B & window=20, std=2 \\
               & ATR (Average True Range) & window=14 \\
    \midrule
    Volume & OBV (On-Balance Volume) & - \\
           & MFI (Money Flow Index) & window=14 \\
           & CMF (Chaikin Money Flow) & window=20 \\
           & Force Index & window=13 \\
    \midrule
    Other & Lagged Close Prices & Lags: 1h, 2h, 3h, 4h, 5h \\
          & Lagged Volume & Lags: 1h, 2h, 3h, 4h, 5h \\
          & Rolling Mean (Close) & Windows: 6h, 12h, 24h \\
          & Rolling Std (Close) & Windows: 6h, 12h, 24h \\
          & High/Low Ratio & High$_t$ / Low$_t$ \\
          & Close/Open Ratio & Close$_t$ / Open$_t$ \\
          & Price Change (1h) & (Close$_t$ - Close$_{t-1}$) / Close$_{t-1}$ \\
          & Price Change (6h) & (Close$_t$ - Close$_{t-6}$) / Close$_{t-6}$ \\
    \bottomrule
  \end{tabular}
\end{table}
\renewcommand{\arraystretch}{1} 

\newpage

\section{Baseline Model Configurations} 
\label{app:baseline_configs_section}
Table~\ref{tab:app_baseline_hyperparams} provides the specific hyperparameters used for each baseline model after tuning on the validation set.

\begin{table}[H]
\centering
\caption{Hyperparameters and Configuration of All Baseline Models (Post-Tuning)}
\label{tab:app_baseline_hyperparams}
\begin{tabular}{lll}
\hline
\textbf{Model} & \textbf{Key Hyperparameters} & \textbf{Value/Setting Chosen} \\
\hline
ARIMA & (p, d, q) & (5, 1, 0) (Order selected via AIC on training data) \\
GARCH & (p, q) & (1, 1) (Commonly used, fit on returns) \\
\midrule
SVM & Kernel & RBF \\
& C (Regularization parameter) & 1.0 \\
& Gamma & 0.1 \\
\midrule
Random Forest & n\_estimators & 100 \\
& max\_depth & 10 \\
& min\_samples\_split & 5 \\
& class\_weight & 'balanced' \\
\midrule
XGBoost & n\_estimators & 100 \\
& max\_depth & 6 \\
& learning\_rate & 0.05 \\
& subsample & 0.8 \\
& colsample\_bytree & 0.8 \\
& scale\_pos\_weight & (Calculated based on training class imbalance) \\
\midrule
LightGBM & n\_estimators & 100 \\
& max\_depth & 6 \\
& learning\_rate & 0.05 \\
& num\_leaves & 31 \\
& boosting\_type & 'gbdt' \\
& class\_weight & 'balanced' \\
\midrule
LSTM & Layers & 3 (Sequential) \\
& Units per layer & [128, 64, 32] \\
& Dropout (after each LSTM layer) & 0.2, 0.3, 0.1 \\
& Batch size & 64 \\
& Optimizer & Adam (learning\_rate=0.001) \\
\midrule
Transformer & Encoder Layers & 2 \\
& Attention Heads (MultiHeadAttention) & 4 \\
& d\_model (Embedding Dimension) & 64 \\
& ff\_dim (Feed-forward dim) & 128 \\
& Dropout rate & 0.1 (attention), 0.2 (feed-forward) \\
& Batch size & 64 \\
& Optimizer & Adam (learning\_rate=0.001) \\
\midrule
CNN-LSTM & 1D Conv Filters & [64, 128] (Sequential) \\
& Kernel Size (1D Conv) & [3, 5] \\
& Activation (Conv) & ReLU \\
& MaxPooling (after Conv) & Pool size = 2 \\
& LSTM Units (after CNN block) & 128 \\
& Dropout (after LSTM) & 0.2 \\
& Batch size & 64 \\
& Optimizer & Adam (learning\_rate=0.001) \\
\hline
\end{tabular}
\end{table}

\newpage

\section{Additional Confusion Matrices}
\label{app:acm}
\begin{figure}[H]
\centering
\includegraphics[width=\linewidth]{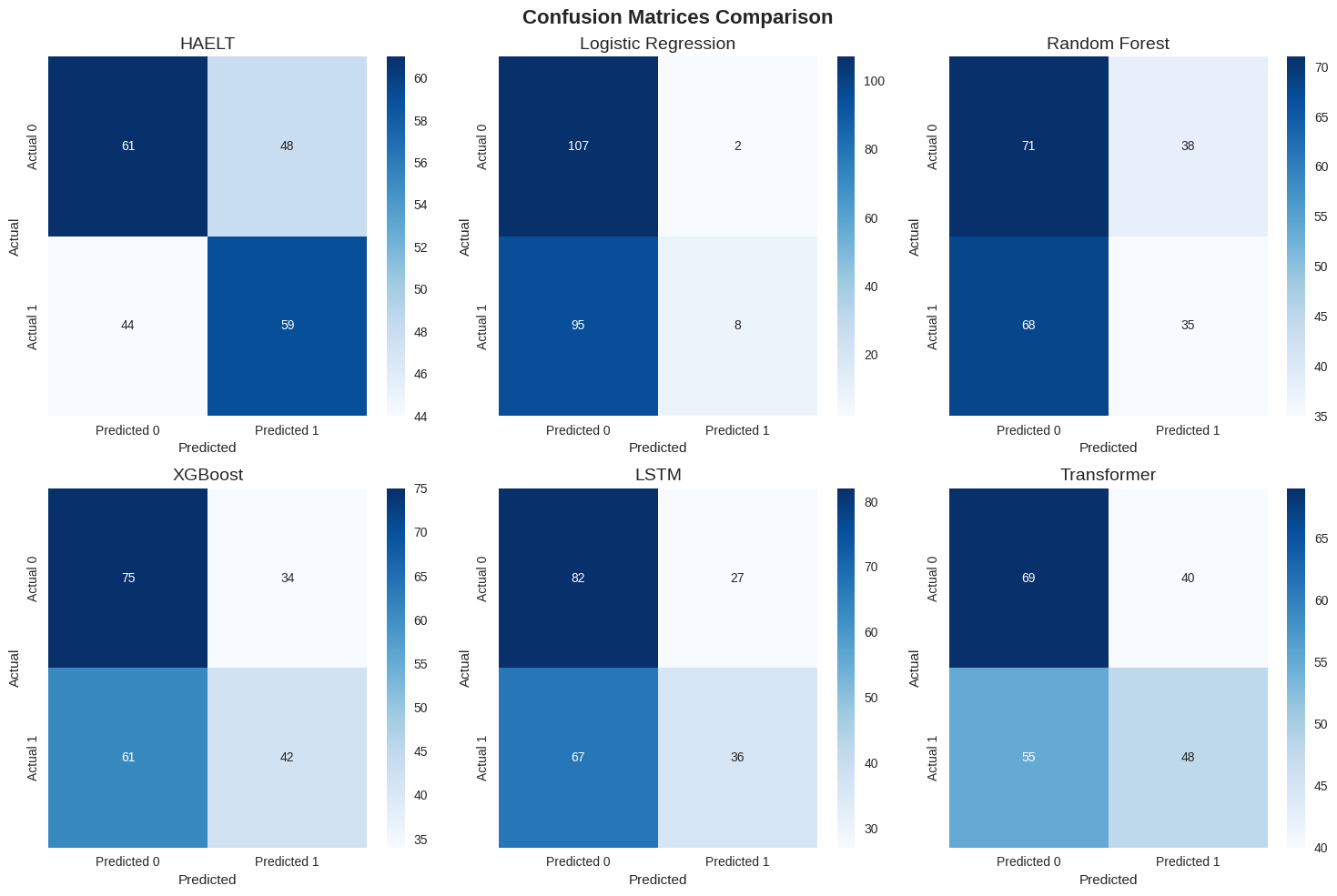} 
\caption{Confusion matrices for key baseline models (e.g., LSTM, Transformer) on the test set.}
\label{fig:app_confusion_baselines}
\end{figure}

Figure~\ref{fig:app_confusion_baselines} shows confusion matrices for key baseline models for comparison with HAELT's matrix in the main text.

\section{Computational Environment}
\label{app:environment_section}
All experiments were conducted on a system with GPU acceleration, characterized by the following specifications:
\begin{itemize}
    \item \textbf{Hardware}: 
    \begin{itemize}
        \item GPU: NVIDIA Tesla P100
        \item CPU: Intel Xeon 
        \item RAM: Approximately 30GB
    \end{itemize}
    \item \textbf{Software}: 
    \begin{itemize}
        \item OS: Linux 
        \item Python Version: 3.11.11
    \end{itemize}
    \item \textbf{Key Libraries}:
    \begin{itemize}
        \item TensorFlow / Keras: 2.12.0
        \item scikit-learn: 1.2.2
        \item pandas: 2.0.3
        \item numpy: 1.23.5
        \item matplotlib: 3.7.1
        \item seaborn: 0.12.2
        \item ta (Technical Analysis Library): 0.10.2
        \item xgboost: 1.7.5
    \end{itemize}
\end{itemize}

\end{document}